%% file: output.tex
\pdfoutput=1 
\documentclass[10pt,twocolumn,letterpaper]{article}
\usepackage{iccv}
\usepackage{times}
\usepackage{epsfig}
\usepackage{graphicx}
\usepackage{enumitem}
\usepackage{amsmath}
\usepackage{amssymb}
\usepackage{booktabs}
\usepackage[accsupp]{axessibility}  

\newcommand{\iccvnew}[1]{{\color{black} #1}}
\newcommand{\smallsec}[1]{\vspace{0.15cm} \noindent {\bf #1.}}

\usepackage[breaklinks=true,bookmarks=false]{hyperref}

\iccvfinalcopy 

\def\iccvPaperID{7337} 

\ificcvfinal\fi

\begin{document}

\title{Gender Artifacts in Visual Datasets}

\author{
\text{Nicole Meister}\textsuperscript{1\dag*},  \ \  \text{Dora Zhao}\textsuperscript{2\dag*},  \ \  \text{Angelina Wang}\textsuperscript{3},  \ \  \text{Vikram V. Ramaswamy}\textsuperscript{3}, \\ 
\ \  \text{Ruth Fong}\textsuperscript{3}, \ \  \text{Olga Russakovsky}\textsuperscript{3}\\
\\
\textsuperscript{1}\text{Stanford University} \ \ \ \  
\textsuperscript{2}\text{Sony AI} \ \ \ \  
\textsuperscript{3}\text{Princeton University} \\
\footnotesize{*Equal contribution \ \ \ \ \dag Work done as a student at Princeton University } \\
{\tt\small \{nmeist, dorothyz\}@stanford.edu}
}

\maketitle
\thispagestyle{empty}
\begin{abstract}
Gender biases are known to exist within large-scale visual datasets and can be reflected or even amplified in downstream models. Many prior works have proposed methods for mitigating gender biases, often by attempting to remove gender expression information from images. To understand the feasibility and practicality of these approaches, we investigate what ``gender artifacts'' exist in large-scale visual datasets. We define a ``gender artifact'' as a visual cue correlated with gender, focusing specifically on cues that are learnable by a modern image classifier and have an interpretable human corollary. Through our analyses, we find that gender artifacts are ubiquitous in the COCO and OpenImages datasets, occurring everywhere from low-level information (e.g., the mean value of the color channels) to higher-level image composition (e.g., pose and location of people). Further, bias mitigation methods that attempt to remove gender actually remove more information from the scene than the person. Given the prevalence of gender artifacts, we claim that attempts to remove these artifacts from such datasets are largely infeasible as certain removed artifacts may be necessary for the downstream task of object recognition. Instead, the responsibility lies with researchers and practitioners to be aware that the distribution of images within datasets is highly gendered and hence develop fairness-aware methods which are robust to these distributional shifts across groups.
\end{abstract}

\section{Introduction}
\input{Input/Introduction_v3}

\section{Related Work}
\input{Input/RelatedWork}

\section{Setup}
\label{sec:method}
\input{Input/Methods}
\section{Resolution and Color}
\input{Input/Experiments}
\section{Person and Background}
\input{Input/PersonScene}
\section{Contextual Objects}
\input{Input/ContextualObj}
\section{Fairness through Blindness Methods}
\input{Input/FairnessBlindness}
\section{Discussion}
\input{Input/Discussion}
\section{Conclusion}
\input{Input/Conclusion}
\begin{flushleft}
\smallsec{Acknowledgements}
This material is based upon work supported by the National Science Foundation under Grant No. 1763642, Grant No. 2112562, Grant No. 2145198, and the Graduate Research Fellowship to A.W., as well as by the Princeton Engineering Project X Fund. Any opinions, findings, and conclusions or recommendations expressed in this material are those of the author(s) and do not necessarily reflect the views of the National Science Foundation.
\end{flushleft}

{\small
\bibliographystyle{ieee_fullname}
\bibliography{ref,visualai}
}
\clearpage
\appendix
\Large{Appendix}
\newline
\newline
\normalsize
\input{Input/Appendix}

\end{document}

%% file: Input/Introduction_v3.tex
It has been well established that machine learning systems contain gender biases~\cite{bolukbasi2016word,caliskan2017semantics,hendricks2018women,wang2019balanced,zhao2021captionbias,mals}. For instance, image tagging systems have labelled similarly posed female politicians differently than their male counterparts~\cite{schwemmer2020diagnosing}, and image search engines can return stereotypical results mirroring harmful gender roles~\cite{kay2015unequal,noble,otterbacher2017competent}.

\begin{figure}[t!]
    \centering
    \includegraphics[width=\linewidth]{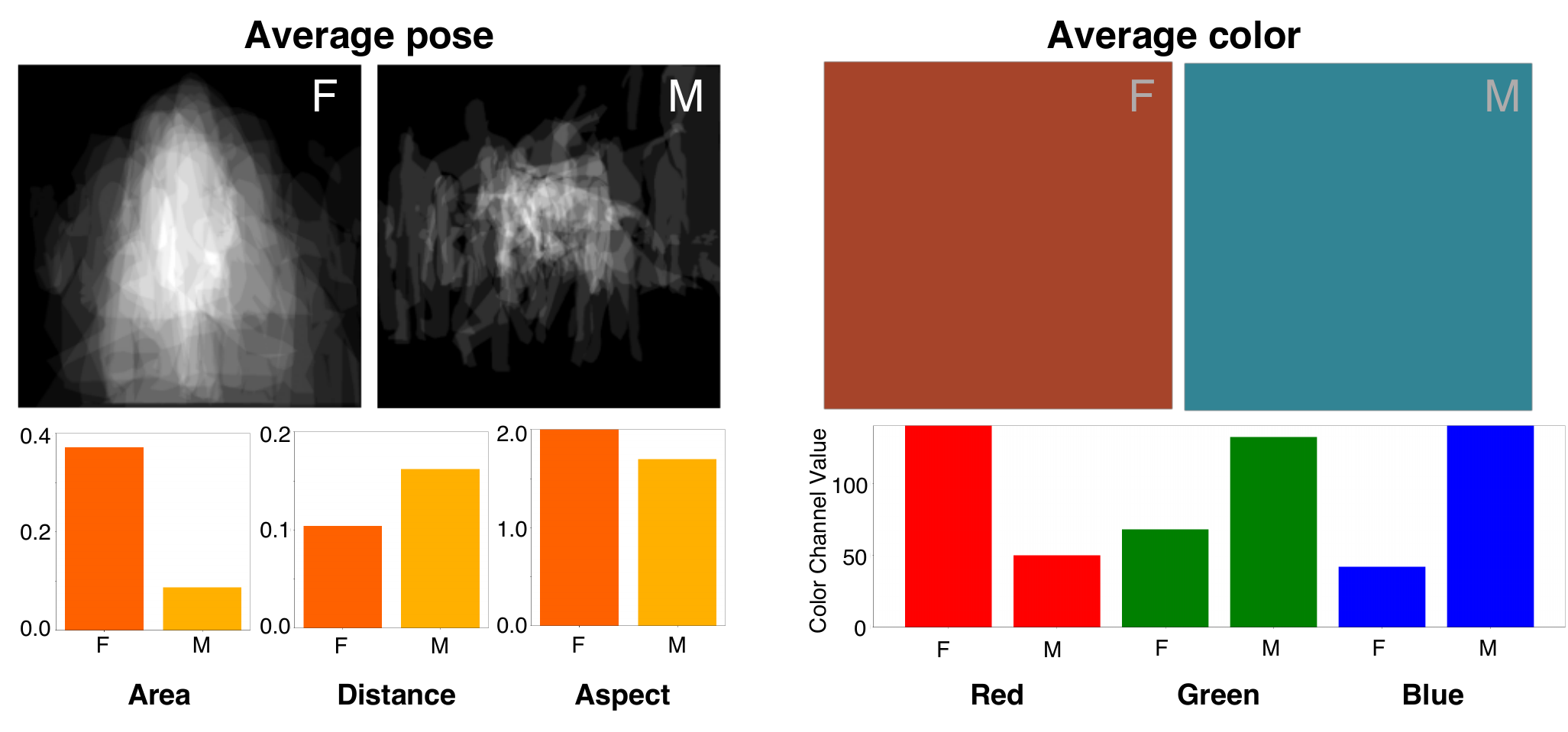}
    \vspace{-0.25in}
    \caption{
    We aim to understand gender artifacts in image datasets by training a classifier to distinguish between images containing a person labelled ``female'' vs. ``male''.
    We find that such a classifier frequently performs significantly above random chance even when trained and evaluated on modified versions of images (e.g. where a person's \emph{appearance} is obscured). 
    On the left, we trained a CNN on COCO~\cite{lin2014microsoft} images with only the person \emph{segmentation mask} visible; on the right, we trained a logistic regression classifier using only 3 features (i.e. \emph{average color}) per COCO image.
    Both classifiers perform significantly above random chance, illustrating how deeply gender artifacts are embedded within the dataset.
    We visualize the classifiers' top 20 (most ``female'') and bottom 20 (most ``male'') images (top), and further show the differences between their segmentation masks (left bottom) in normalized area, distance from the center, and aspect ratio, and the differences between the R, G and B features (right bottom).
    These results suggest that gender artifacts are pervasive in computer vision datasets, and current ``fairness through blindness'' methods that aim to ``remove'' gendered information may not be as effective as hoped.
    } 
    \label{fig:fig1}
\end{figure}

A prevailing assumption in the fairness community is that biases in models originate in biases in the input data. Thus, there has been an effort to mitigate these dataset biases (and thus also model biases). While many biases stem from the lack of representation of certain demographic groups, including women~\cite{buolamwini2018gender,zhao2021captionbias,mals}, naively balancing gender distributions has been shown to be insufficient for mitigating model biases~\cite{wang2019balanced}. As a result, a large vein of work has attempted to pursue ``fairness through blindness''~\cite{dwork2011awareness} which operates under the assumption that removing gendered information can remove the potential for gender bias in the dataset. These attempts include using segmentation masks to occlude the person in images~\cite{bhargava2019caption,hendricks2018women,tang2021mitigating} or removing image features correlated with gender~\cite{wang2021gender,wang2019balanced}.   

In our work, we focus not on mitigating these dataset biases, but rather on exploring to what extent gendered information can truly be removed from the dataset. To do so, we develop a framework to identify \emph{gender artifacts}, or visual cues correlated with gender labels. We use the term ``artifact'' to emphasize how these correlations may not be related to gender presentation, rather simply a result of the dataset collection process. While there can be infinitely many potential correlations in an image, we focus on those that are learnable (i.e., result in a learnable difference in a model) and interpretable (i.e., have an interpretable human corollary, such as color and shape).\footnote{See Appendix~\ref{sec:artifact_constraints_appendix} for details on artifact constraints.} Overall, we do not claim (nor believe) the existence of a specific gender artifact necessarily leads to disparate performance on downstream tasks for different groups (a common measure of fairness\cite{buolamwini2018gender,schumann2021step,zemel_fairaware}), however, the \textit{source} of disparate performance lies in gender artifacts which are critical to study. 

To discover gender artifacts, we use a gender classifier, which we refer to as a ``gender artifact model,'' since the model is not predicting gender so much as it is predicting correlations to gender expression in the training dataset. Our goal is to understand what the model uses as predictive features and how variations of the dataset with particular gender artifacts removed may affect its performance. Thus, we systematically manipulate the datasets (e.g., occlude the person, occlude the background, etc.) and train our model on this data to identify potential gender artifacts. This framework for discovering gender artifacts is more complex than prior work~\cite{wang2020revise}, as it goes beyond analyzing annotated attributes to include higher level perceptual components of the image (e.g., resolution, color) and person and background artifacts to gain a deeper understanding of where gender artifacts exist in visual datasets. 

It is important to note we do not condone the use of automated gender recognition in practice. We perform gender classification only to understand differences in data distributions, \textit{not} for the sake of classifying gender itself. We never use the model outputs as the end goal and no causal claims about the observed correlation between an attribute and gender labels can be made. Instead, the outputs are a means for better understanding the data. Furthermore, our use of the term ``gender'' when referencing people in image datasets refers to the binarized perceived gender expression. Since most image datasets do not include self-reported gender identity, we rely on external annotators' perception as gender labels~\cite{schumann2021step}. To be consistent with prior works~\cite{buolamwini2018gender,hendricks2018women,wang2019balanced,zhao2021captionbias,mals}, we use ``male'' and ``female'' to refer to gender expression. Finally, we do not make any normative claims about the gender artifacts (i.e., if it is good or bad); we focus simply on identifying which artifacts exist.

Using our proposed framework, we perform an in-depth analysis into understanding the wide variety of gender artifacts present in two popular image datasets: Common Objects in Context (COCO)~\cite{lin2014microsoft} and OpenImages~\cite{kuznetsova2020open}. Our results show that gender artifacts are \textit{everywhere}, from the shape of the person segmentation mask (Fig.~\ref{fig:fig1}) to randomly selected contextual objects. 
Even the \emph{average color} of the image is sufficient to distinguish between the genders: when each image is represented by just three features, the mean pixel value of the red, green, and blue color channels, the classifier is able to achieve an AUC of $58.0\% \pm 0.4$ in COCO and $59.1\% \pm 0.4$ in OpenImages.

These findings have the following implications: 
\begin{itemize}[]
    \item Many prior works have proposed mitigation methods that attempt to remove gender expression information from the image~\cite{alvi,bhargava2019caption,hendricks2018women,tang2021mitigating,wang2019balanced}. In contrast, we show \textbf{gender artifacts are so intricately embedded in visual datasets that attempts to remove them via mitigation techniques may be a futile endeavor}. This is evidenced by experiments where even after a person is \emph{entirely occluded} with a rectangular mask and the background is visible, the gender artifact model is nevertheless able to reliably distinguish which of the two genders is present in the image, achieving an AUC of $70.8\%$ in COCO~\cite{lin2014microsoft} and $63.0\%$ in OpenImages~\cite{kuznetsova2020open}. \iccvnew{Using a popular adversarial approach~\cite{wang2019balanced} as a case study, we find $76.8\%$ of the removed information comes from the scene rather than the person.}
    \item We more realistically advise practitioners to \textbf{adopt \emph{fairness-aware} models}~\cite{dwork2011awareness,zemel_fairaware} and disaggregated evaluation metrics~\cite{barocas2021disagg}, which explicitly account for potential discrepancies between protected groups. 
    \item Since there are so many salient artifacts in image datasets that are (spuriously) correlated with gender, our findings point to an \textbf{incoherence of gender prediction}. Any time a model predicts ``gender,'' we should wonder what the prediction refers to; it could easily be relying on the background scene color, rather than societally meaningful forms of gender expression.
\end{itemize}

%% file: Input/RelatedWork.tex
\smallsec{Understanding model biases} 
Understanding where biases arise in models is important to inform mitigation strategies. One proposed method is experimentally manipulating features in an image to isolate sources of gender biases. This has been done using simple image processing techniques, such as changing the image brightness~\cite{muthukumar2018understanding}, manually finding counterfactual examples~\cite{stock2018convnets}, or using generative adversarial networks (GANs) to synthetically manipulate attributes~\cite{balakrishnan2020towardsCB,denton2019detecting,ramaswamy2020debiasing,sharmanska2020contrastive}. Another approach is to use interpretability methods, such as attention heatmaps, to understand where the model focuses when making a prediction~\cite{hendricks2018women}. While we use classifiers as a tool for understanding gender artifacts, we focus on how these gender artifacts manifest in \textit{datasets} rather than in the models. 

\smallsec{Identifying dataset biases}
The presence of bias in datasets has been well-studied. While dataset bias can be analyzed along many axes both with respect to demographic~\cite{buolamwini2018gender,park2021understanding,wang2020revise,zhao2021captionbias} and non-demographic attributes~\cite{de2019does,torralba2011unbiased,wang2020revise}, we focus on the presence of social biases in visual datasets. Image datasets have been found to include imbalanced demographic representation~\cite{buolamwini2018gender,yang2020fair,zhao2021captionbias}, stereotypical portrayals~\cite{caliskan2017semantics,schwemmer2020diagnosing,van2016stereotyping}, and even harmful or offensive content~\cite{prabhu2020large,zhao2021captionbias}. Looking more closely into gender artifacts, prior works have considered differences with respect to demographic attributes~\cite{prabhu2020large,buolamwini2018gender,zhao2021captionbias} and contextual objects~\cite{prabhu2020large,wang2020revise,mals} (e.g., instance counts, distance from objects). 

\smallsec{Mitigating dataset and model biases}
Prior works have proposed interventions at both the dataset and algorithmic level. Proposed techniques for mitigating dataset bias include manual data cleanup~\cite{yang2020fair} and applications of synthetic methods, such as GANs~\cite{ramaswamy2020debiasing,sattigeri2019fairness,sharmanska2020contrastive}. Algorithmic interventions have also been proposed to mitigate gender biases. One common approach is to remove gendered information from the image by blurring or occluding pixels corresponding to people ~\cite{alvi,bhargava2019caption,hendricks2018women,tang2021mitigating,wang2019balanced}. For example, Hendricks et al.~\cite{hendricks2018women} developed techniques encouraging the model to be ``confused'' when predicting gendered words if the person is occluded, and Wang et al.~\cite{wang2019balanced} use adversarial methods that learn to obscure image features that correspond with gender. Other examples of algorithmic mitigation techniques include domain independent training~\cite{wang2020fair} and corpus-level constraints~\cite{mals}. \iccvnew{In contrast, we focus on analyzing where gender artifacts may \emph{originate} in visual datasets. We evaluate what these interventions that remove gender artifacts actually address, and in turn, which artifacts may still remain even after applying mitigation techniques.}

\smallsec{Analysis of data collection practices}
Analyzing dataset collection practices is key to better understanding where and how dataset biases arise; see e.g., Paullada et al.~\cite{paullada2021data} for an overview of how collection, annotation, and documentation practices introduce biases into visual datasets. For example, the common practice of scraping images from the internet injects biases inherently present in image search engines, such as gender~\cite{kay2015unequal,noble} or geographic biases~\cite{de2019does,shankar2017no}. In addition, prior work~\cite{otterbacher2019we} has identified that crowdsourced ground truth labels can be a potential source of bias as annotators may systematically differ in their perceptions, often influenced by their demographic background.  
In response, fairer dataset collection practices have been suggested, e.g., drawing inspiration from other disciplines~\cite{hutchinson2021towards,jo2020lessons}. One focus has been on improving transparency, including creating public datasheets and tools to guide researcher intervention~\cite{prabhu2020large,gebru2021datasheets,holland2018dataset}. There has also been an effort to create more structured processes, including creating checklists and advocating for more institutional oversight~\cite{madaio2020co,peng2021mitigating}.

%% file: Input/Methods.tex
\subsection{Datasets} We focus our analysis on COCO~\cite{lin2014microsoft} and OpenImages~\cite{kuznetsova2020open} as they are widely used across different computer vision tasks~\cite{ferraro2015survey,hossain2019comprehensive,minaee2021image}, making it particularly important to investigate biases present in these datasets~\cite{creel2021leviathan}. COCO has been the testbed for previous work on bias mitigation~\cite{hendricks2018women,wang2019balanced,mals} making it central to our analyses. We use OpenImages to validate that our findings on COCO are generalizable. These datasets are also unique in that we are able to derive labels from existing annotations for perceived gender expression to train a gender artifact model. 

For COCO, following  Zhao et al.~\cite{mals}, we derive gender labels using the COCO 2014 captions as this practice is standard in existing bias mitigation methods~\cite{hendricks2018women,wang2019balanced,mals}.
For OpenImages, we use the More Inclusive Annotations for People (MIAP)~\cite{schumann2021step} and exclude images with more than one person of different gender labels. Notably both datasets are skewed male: 69.2\% COCO and 62.9\% OpenImages.

\subsection{Model} Our gender artifact model is a ResNet-50~\cite{He2015} pre-trained on ImageNet~\cite{imagenet_cvpr09} (see Appendix~\ref{sec:pretrain_appendix} for details). To remove gender artifacts, we train and evaluate on selectively modified versions of images. For a threshold agnostic performance measure, we report area under the ROC curve (AUC) and standard error on the test set. An AUC of $50\%$ suggests the classifier cannot distinguish between male and female gender labels (i.e., there are no image artifacts that allow the classifier to distinguish between these labels).

To note, AUC is not a fairness metric in the way equalized odds~\cite{Moritz}, equality of opportunity~\cite{Moritz}, or statistical/prediction parity~\cite{Barocas2018FairnessAM} are. A higher or lower AUC does not indicate that a dataset is more or less fair. Rather, we use AUC to help understand where correlations with gender arise in visual datasets. We follow similar approaches to research within the computational biology community that finds that gender, race, and age can be inferred from various medical imaging techniques (e.g., chest and hand X-rays, mammograms, retinal fundus photography) due to subtle cues in the input space~\cite{ReadingRace,howard,Poplin}. They similarly predict an attribute from medical images and report AUC.

\subsection{Ethical considerations}
While we only use gender expression classification as a discovery mechanism for visual artifacts, we acknowledge that our setup, which predicts ``male'' or ``female'', reifies the notion of binary gender and grants it legitimacy. Gender classifiers are fundamentally imperfect as gender is reduced to a simplistic binary that can be harmful to individuals from the trans and/or non-binary community who may not fit into these narrow categories \cite{hamidi2018gender,keyes2018agr}. While it is important to understand where gender biases may arise in automated systems, we do not condone the use of automated gender recognition in practice. The purpose of a gender classifier in this project is solely for the study of gender bias propagation. We are aware that by simply calling a gender classifier a ``gender artifact model'' does not change what it fundamentally is --- however, by naming it as such we underscore its purpose as a method of studying gender artifacts \textit{only} and not for its prediction output of gender.

%% file: Input/Experiments.tex
\begin{figure}[]
    \centering
    \includegraphics[width=\linewidth]{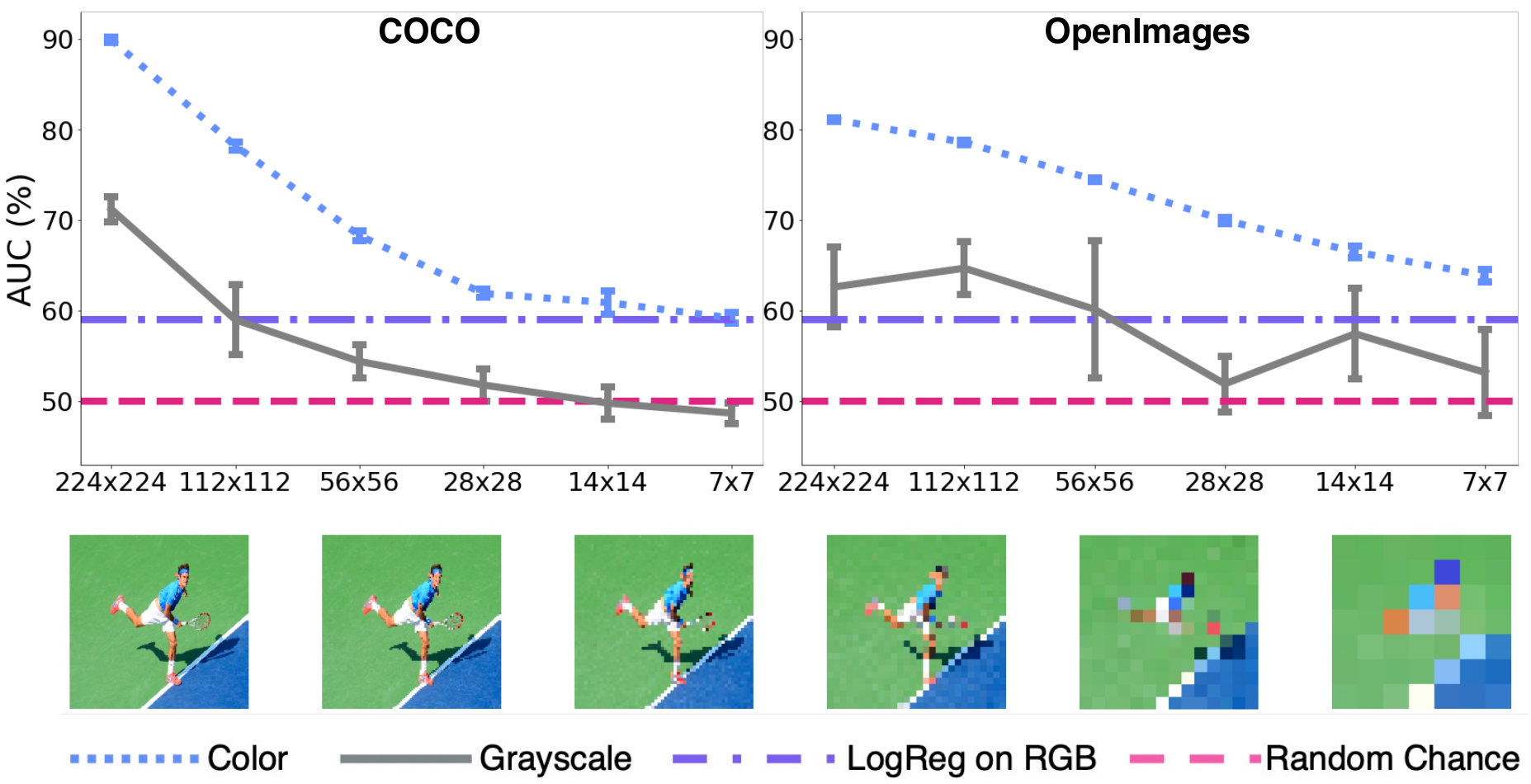}
    \caption{AUC of the gender artifact models on the COCO~\cite{lin2014microsoft} (top left) and OpenImages~\cite{kuznetsova2020open} (top right) datasets. The models trained and evaluated on color images (blue curves) perform significantly above random chance (pink lines) even as the image resolution gets as low as 7x7. The models trained on grayscale images (gray curves) approach random chance around 28x28 resolution. A sample image at different resolutions is shown on the bottom. Most interestingly, as also highlighted in Fig.~\ref{fig:fig1}, the logistic regression classifier trained on just the mean red, green and blue channel values (purple line) performs above random chance on both datasets,} underscoring the role of color as a gender artifact and the overall ubiquity of gender artifacts in image datasets.
    \label{fig:color_res}
    
\end{figure}

\label{sec:rescolor}
To start, we analyze the higher-level perceptual components of the image that may serve as gender artifacts. We specifically investigate the resolution and color of the images. 
The scenarios we experiment with are images (originally 224x224) downsampled to 112x112, 56x56, 28x28, 14x14, 7x7. As the images decrease in resolution, it becomes harder to distinguish the contents, such that color becomes one of the more salient artifacts. Thus, we also consider these images of lower resolution in grayscale, to better understand what the model is relying on when the image has been distorted such that individual objects and people are not perceptible. All results are in Fig. \ref{fig:color_res}.

The experiments in manipulating image resolution are motivated by previous work that finds a spatial resolution of 32x32 is sufficient to identify the semantic category of real world scenes~\cite{torralba_2009}. We downsample the images to analyze what happens to a gender artifact model when the input images are beyond a discernible image resolution. The model's AUC plateaus at around 61.9\% for color and 51.9\% (near-random) for grayscale COCO images at 28x28 resolution. These results suggest that the shapes in an image are no longer meaningful at that resolution, and the gender artifact model must be predominantly relying on color features.

It is also surprising to see that a gender artifact model trained and evaluated on 7x7 images (a resolution at which almost all image artifacts except color are removed from an image) still achieves an AUC of above random chance ($59.2$\% for COCO and $63.9$\% for OpenImages).

Exploring this further, we train a logistic regression model\footnote{This model is trained with L2 regularization, $\lambda=1$.} on three features: the mean pixel value of the red, green, and blue color channels. Surprisingly, this simple model achieves an AUC of 58.0\% $\pm$ 0.4 on COCO and 59.1\% $\pm$ 0.4 on OpenImages. This likely occurs as many images of people labeled male are in grassy sports fields, whereas many images of people labeled female are of the person, generally of lighter skin tone~\cite{zhao2021captionbias}, up close. The fact that just three color values are sufficient for a model to differentiate between images of two different genders, emphasizes how ubiquitous gender artifacts are in the dataset. 

%% file: Input/PersonScene.tex
\label{sec:personscene}
Next, we turn to disentangling gender artifacts arising from the person versus the background (i.e., non-person parts of the image). Prior bias mitigation works~\cite{hendricks2018women,wang2019balanced} assume that background information should not contain gender artifacts. For example, Hendricks et al.~\cite{hendricks2018women} propose a bias mitigation strategy for image captioning that encourages the model to be ``confused'' when predicting gendered words if the person is occluded. Similarly, adversarial methods~\cite{wang2019balanced} learn to obscure person information from the input image. However, additional work has shown models can achieve non-trivial accuracy in image classification by relying on the background alone~\cite{xiao2021noise}. A natural question is whether other aspects of the image (e.g., background, person pose, person size) contain gender artifacts as well.

\begin{figure}[t!]
    \centering
    \includegraphics[width=\linewidth]{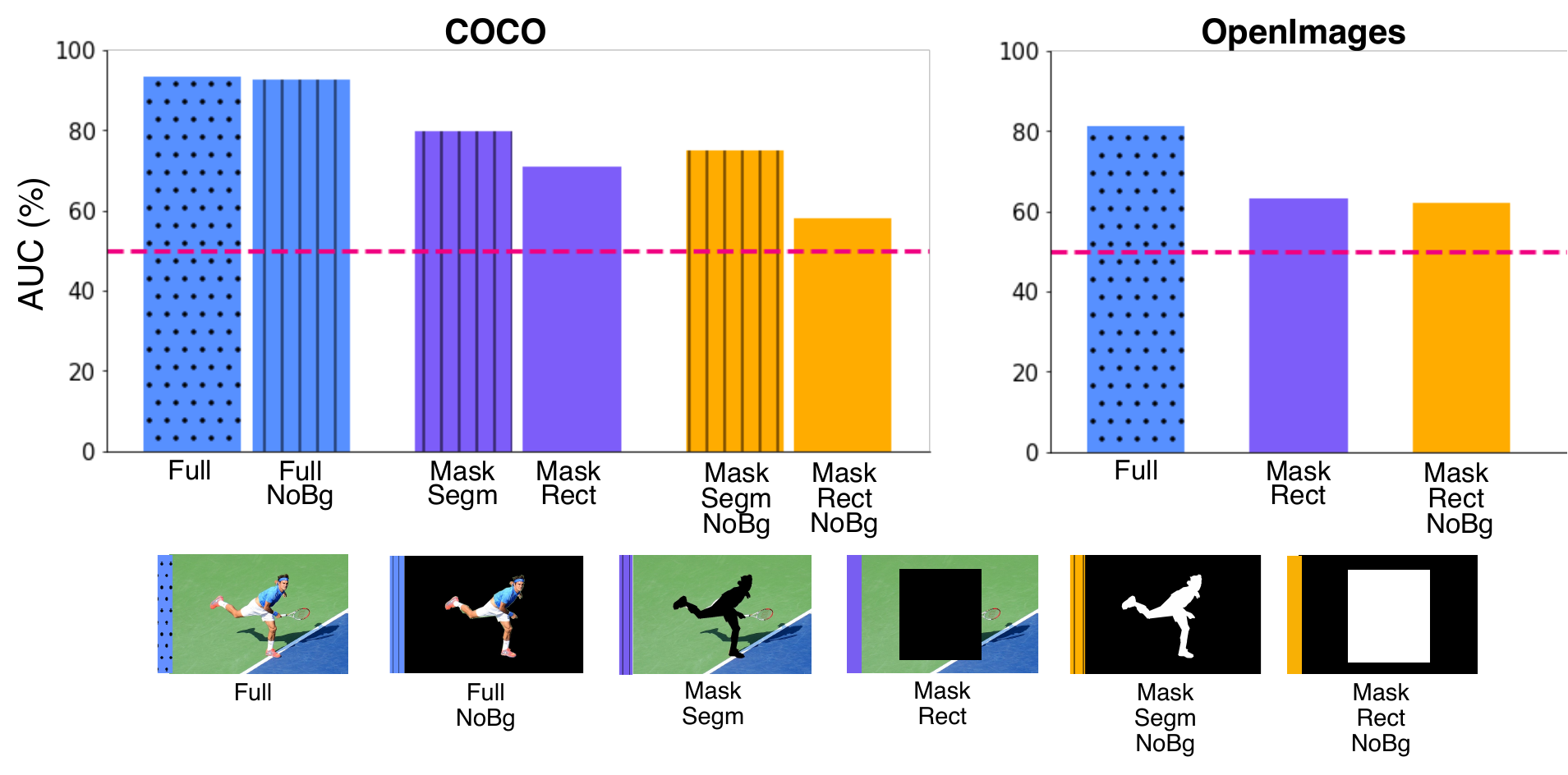}
    \caption{AUC of gender artifact models (top) trained and evaluated on images with different levels of occlusions (shown on bottom). The naming refers to what from the person is shown in an image: all person pixels  (Full),  segmentation mask (MaskSegm), or bounding box (MaskRect). ``NoBg'' 
    denotes a masked background. 
    The pink line represents random chance. 
} 
    \label{fig:sec4}
\end{figure}

\smallsec{Experimental setup} 
We use a series of occlusions on COCO and OpenImages. For simplicity, our experiments adhere to the nomenclature shown at the bottom of Fig.~\ref{fig:sec4}.

We train and evaluate the gender artifact model from Sec.~\ref{sec:method} on the manipulated images. All models are trained and tested on the same occlusions (e.g., model trained on images with occluded background are tested on images with occluded background). Additionally, since COCO and OpenImages are skewed by gender, we consider the effect of the imbalanced datasets, replicate these experiments on balanced dataset, and find similar results (see Appendix~\ref{sec:balance_appendix}).

\smallsec{Person's appearance is a gender artifact} We start by examining the artifacts arising from the person pixels. First, we consider the model trained on Full (i.e., no occlusions) which achieves an AUC of $93.4\% \pm 0.2$ on COCO. We then compare the performance on Full with on Full NoBg (i.e., only the outline of the person is visible against a black background).\footnote{Results when occluding with white pixels are similar; in Appendix~\ref{sec:person_scene_appendix}.} The model's performance does not change considerably even after the background is removed, achieving an AUC of $92.7\% \pm 0.2$ (see blue bars in Fig.~\ref{fig:sec4}).

\smallsec{A gender artifact model performs above random chance when trained on the person’s shape and location} Subsequently, we question \textit{what} about the person, namely in their shape and location, is a gender artifact. To isolate artifacts arising from the person's shape, we occlude the background and then gradually remove information about the person's appearance (see orange bars in Fig.~\ref{fig:sec4}). First, looking at MaskSegm NoBg (i.e., person occluded with white pixels against the black background), we observe that the model performs considerably better than random chance with an AUC of $74.8\% \pm 0.3$ for COCO. This result may be expected as the segmentation shapes themselves still reveal information about appearance, such as clothing or hair length. 

Next, removing even more information about the person, we train our model using only the 17 person keypoint locations from COCO.

The model continues to perform above random chance (AUC of $64.8\% \pm 0.4$). Further, to disentangle the role of pose versus other image factors (e.g., size, location), we resize the keypoints to have an area of 4,000 pixels and center the keypoints in the image. These images are then used to train a multi-layer perceptron (MLP) for 100 iterations, achieving an AUC of $58.0\% \pm 0.0$, suggesting pose is a salient gender artifact. Finally, we inspect the poses with the greatest absolute scores and find the poses predicted to be more likely male are smaller and in action (e.g., playing a sport, jumping) whereas those predicted to be more likely female tended to be larger and standing still.

\smallsec{There is a learnable difference between the size and location of people of different gender expressions} Since we observe a difference in the size for poses, we further question whether the model can use only the size and location of people as gender artifacts. We use MaskRect NoBg, (i.e., only the bounding box around the person occluded with white pixels against a black background). Although performance decreases from when pose information was provided, the model continues to find a learnable difference with an AUC of $58.0\% \pm 0.4$ on COCO. This pattern is also reflected in OpenImages ($62.2\% \pm 0.4$). In fact, in the COCO training set, the size of people labeled as female (normalized by the image area) is $0.18 \pm 10^{-4}$ and $0.13 \pm 10^{-5}$ for people labeled as male. This difference is statistically significant with $p< 0.01$ level. The same pattern holds for OpenImages. 

We also verify whether location is predictive. Using a one-hot encoding of the 32x32 image where each pixel is set to 0 except the center of the person bounding box, we train an MLP for 100 iterations. The model achieves an AUC of $51.5\% \pm 0.0$. Considering \textit{only} the center pixel's location provides a slight learnable difference.

\smallsec{Backgrounds are significantly different for images with people of different gender expressions}

Next we occlude information about the person but keep the background intact (see purple bars in Fig.~\ref{fig:sec4}). 

For both MaskSegm ($79.6\% \pm 0.3$) and MaskRect ($70.8\% \pm 0.3$) the model performs better than random chance on COCO. The same is true for MaskRect for OpenImages ($63.1\% \pm 0.4$). Here, our occlusions match those from proposed mitigation strategies~\cite{hendricks2018women,wang2019balanced}. Once again, occluding only the person's appearance is insufficient to remove all the gender artifacts in the image. Thus, any attempt to remove person-related gender artifacts shifts the source of artifact information from the person to the background, thus not satisfying the goal of removing these artifacts altogether.

\begin{figure}[t]
\centering
\includegraphics[width=\linewidth]{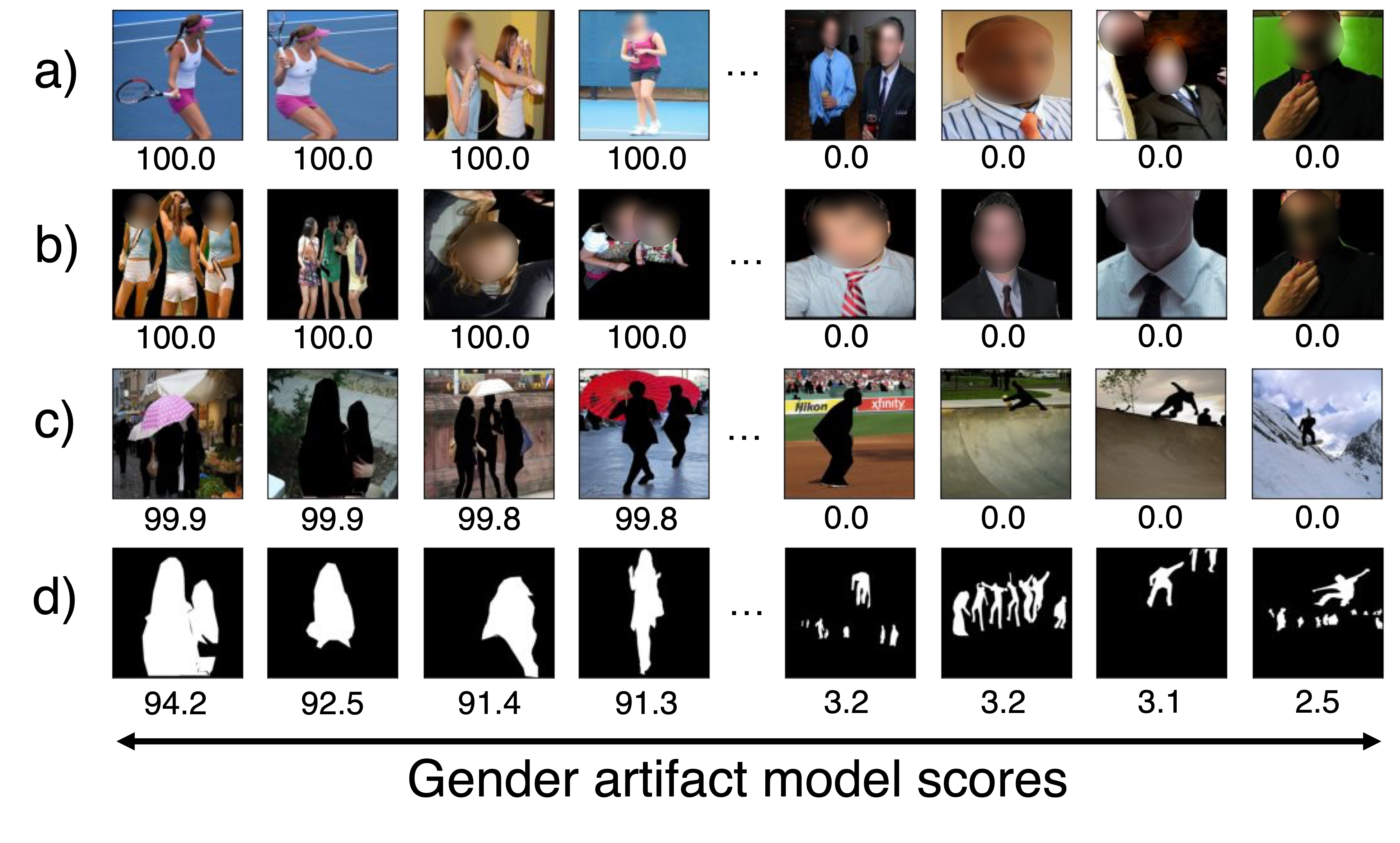}
\caption{Images with highest and lowest scores for models trained on the following occlusions: a) Full, b) Full NoBg, c) MaskSegm, and d) MaskSegm NoBg. 100 corresponds to the model predicting ``female '' and 0 to ``male.''  Face pixelation is not in the COCO image but included to partially preserve privacy.}
\label{fig:qualanalysis}
\end{figure}

\smallsec{Qualitative analysis validates our findings} To better understand the impact of background as a gender artifact, we qualitatively inspect images with the highest and lowest scores from our gender artifact model (Fig.~\ref{fig:qualanalysis}). In (c) MaskSegm, the person's location, together with background, is used as a gender artifact; the model associates city backgrounds with ``female'' and athletic scenes with ``male.'' Also, from (d) MaskSegm NoBg, we corroborate our finding that the model learns to associate larger, static poses with ``female'' and smaller, active poses with ``male.'' However, we do note that when the person pixels are included, the model is able to make more confident predictions for images with non-stereotypical settings (e.g., person labeled female playing tennis, person labeled male in a static, close-up shot) as seen in rows (a) and (b). 

\smallsec{Gender artifacts generalize across datasets} 
Finally, we examine whether gender artifacts are dataset-specific, i.e., resulting from a dataset's specific collection process and distribution. We conduct out-of-distribution evaluation (e.g., training on COCO and evaluating on the same occlusion setting on OpenImages). Across occlusion settings, the models perform above random chance even when evaluated on an out-of-distribution test set (Tbl.~\ref{tab:ood}). This suggests that these gender artifacts may be common across visual datasets. To verify that our findings are not restricted to models pre-trained on ImageNet, for which there are known gender biases~\cite{prabhu2020large,yang2020fair}, we also use a model pre-trained on other datasets and obtain similar results (see Appendix~\ref{sec:pretrain_appendix}). 

\begin{table}[t!]
    \footnotesize
    \caption{Out-of-distribution evaluation between the two datasets (e.g., training on COCO and evaluating on OpenImages) and the AUC for the three occlusion settings. We report the $95\%$ confidence interval from bootstrapping.}
    \centering
    \begin{tabular}{p{1.75cm}p{1.2cm}p{1.2cm}p{2cm}}
    \toprule
         Train / Test & Full & MaskRect & MaskRect NoBg \\
    \midrule
         COCO / COCO & $93.4 \pm 0.1$ & $70.7 \pm 0.1$ & $58.3 \pm 0.2$ \\ 
         COCO / OI & $91.2 \pm 0.2$ & $60.7 \pm 0.4$ & $63.2 \pm 0.4$\\
         OI / COCO & $76.4 \pm 0.3$ & $57.9 \pm 0.4$ & $54.4 \pm 0.4$ \\
         OI / OI & $81.2 \pm 0.2$ &  $63.3 \pm 0.4$  & $59.7 \pm 0.7$ \\ 

    \bottomrule 
    \end{tabular}
    \label{tab:ood}
\end{table}

%% file: Input/ContextualObj.tex
\label{sec:contextualobjs} In the previous section, we discovered that gender artifacts are present in both person and background pixels. In this section, we work to better understand these gender artifacts in the background, or non-person parts of the image, and distinguish between two components of the background: the objects\footnote{Albeit limited to the objects labeled by COCO and Open Images.} and the ``scenes'' (i.e., the rest). 

Prior works~\cite{hendricks2018women,wang2019balanced} often assume gender artifacts in the background (i.e., objects or scene) are forms of ``bias'' and aim to mitigate them by studying object-gender co-occurrences. Additionally, prior works such as Singh et al.~\cite{Singh_2020_CVPR} point out the dangers of bias from contextual objects, particularly in visual recognition datasets, and propose mitigation techniques to recognize an object or attribute in the absence of its typical context and account for imbalances in object co-occurrences. 

In this section, we describe how we identify the relative gender artifact contributions of the objects, as well as which objects contribute to the model's ability to differentiate images of people of different genders. 

\smallsec{Top objects contributing to gender prediction} We analyze the role of specific objects in gender prediction by training a logistic regression model using a binary presence-absence vector as input, where each value of the vector corresponds to the presence of an annotated object (1 if the object is present and 0 otherwise).\footnote{Trained with L1 regularization, reg. strength of 1,} and liblinear solver. COCO contains 80 objects chosen by the dataset creators as a representative set of all categories, relevant to practical applications, and occurring with high enough frequency~\cite{lin2014microsoft}. OpenImages contains 599 objects, but explicitly gendered objects (i.e., \texttt{woman}, \texttt{man}, \texttt{girl}, \texttt{boy}) are removed before training the classifier~\cite{kuznetsova2020open}.

\begin{figure}[t!]
\centering
\includegraphics[width=\linewidth]{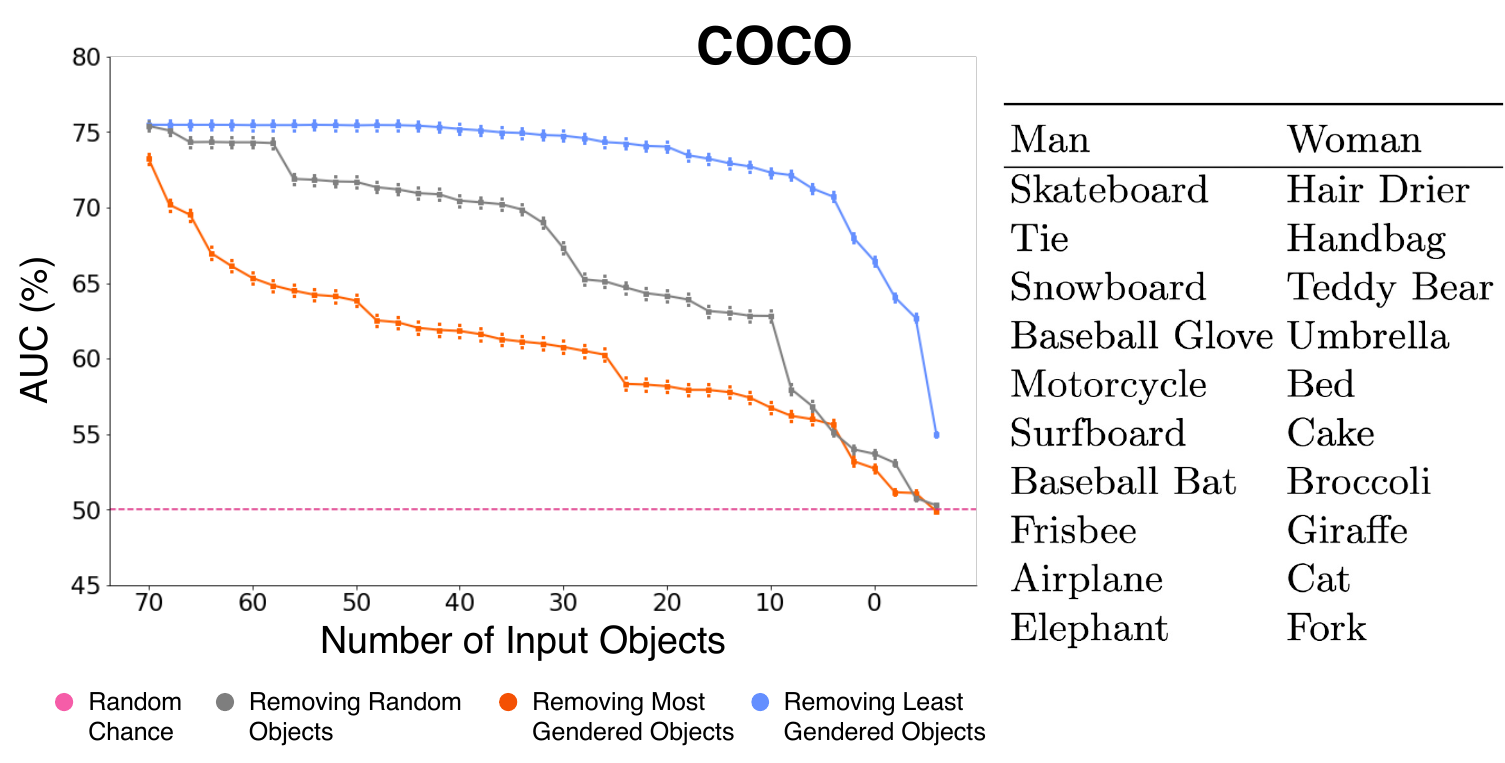}
\caption{
The change in AUC (\%) as objects are iteratively removed from a logistic regression classifier trained on COCO (left). Our goal is to see how many objects must be removed before the classifier performs at random chance. When removing random objects, we report the mean and confidence interval from bootstrapping and train five separate random classifiers (see Appendix~\ref{sec:contextual_obj_appendix}). We also show the 10 most relevant objects in descending order, as identified by the weights of the logistic regression classifier (right).}
\label{fig:logreg_removeobjs}
\end{figure}

The model achieves an AUC of $75.4\%$ on COCO (Fig.~\ref{fig:logreg_removeobjs}) and $63.3\%$ on OpenImages. 
The performance of the OpenImages classifier is significantly lower than that of COCO; this may be attributable to the difference in annotated objects in the train and test distribution (the OpenImages test set has an average of 7.7 object instances annotated per image while the training set has an average of 3.7).\footnote{OpenImages dataset creators generated denser labels for the validation and test splits than for the train split~\cite{kuznetsova2020open}.} 

This classifier’s weights loosely correspond to the object’s role in gender classification. We use the weights to identify the most relevant objects that contribute to gender prediction sorted by importance. For COCO, the \textit{bottom} 10 classifier weights correspond to the top 10 objects most useful to classify male (e.g., \texttt{skateboard}, \texttt{tie}, \texttt{snowboard}). The \textit{top} 10 classifier weights correspond to the top 10 objects most useful to classify female (e.g., \texttt{hair drier}, \texttt{handbag}, \texttt{teddy bear}) (see Fig. \ref{fig:logreg_removeobjs}). 

To further understand the role of contextual objects, we iteratively remove objects from the logistic regression classifier (by decreasing the input vector dimension) to ``break" the classifier. That is, we want to see how many objects need to be removed  before the classifier performs at random chance. We remove the ``most gendered objects" as defined by the object-based logistic regression model's weights as shown in the table in Fig.~\ref{fig:logreg_removeobjs}. In the first iteration, the objects corresponding to the highest and lowest classifier weights are removed and the model is retrained. For the second iteration, the objects corresponding to the top 2 highest and top 2 lowest weights are removed, etc.

Interestingly, while removing the gendered objects does force the AUC to decrease faster, we find that this classifier still performs above 50\% AUC and can achieve an AUC of $72.1\%$ when given 14 of the most gendered objects (as noted by iteratively removing the ``least gendered objects" or objects corresponding to the lowest classifier weights). As shown in Fig. \ref{fig:logreg_removeobjs}, we also experiment with removing randomly selected objects and find that even when given a small number of random objects, the gender artifact model is able to perform above random chance with an AUC of $66.8\%$ trained only the one-hot encodings of 30 randomly chosen objects among the 80 annotated objects and $58.0\%$ with 10 random objects.

The analysis of contextual object underscores the main takeaway: gender artifacts are everywhere. Even when only presented with one-hot encodings of objects, a gender artifact model can still perform above random chance.

%% file: Input/FairnessBlindness.tex
One common approach to fairness in computer vision is that of ``fairness through blindness" where the classifier explicitly does not use the sensitive attribute e.g., gender. Such bias mitigation methods 
will remove gender artifacts and occlude gender-related features~\cite{holland2018dataset,torralba2011unbiased,xiao2021noise} (e.g., face, person segmentation mask, person bounding box~\cite{xiao2021noise}) to encourage the model to learn representations not reliant on gendered information. In this section, we use Wang et al.'s~\cite{wang2019balanced} adversarial approach as a case study and investigate what gender artifacts are being removed or debiased. 

\smallsec{Method}
As in~\cite{wang2019balanced}, we train a U-Net~\cite{ronneberger2015u} on COCO training images to do gender debiasing by removing image pixels corresponding to salient gender artifacts. We then identify and analyze the pixels in COCO validation images (with $\geq 1$ \texttt{person} object) that were removed in the debiasing process (see Appendix~\ref{sec:fair_blind_appendix} for details).

\smallsec{Results} 
Intuitively, we would assume that fairness-through-blindness approaches remove gender artifacts by removing the person; however, in practice, we find the method removes artifacts that overwhelmingly come from the background, not the person. Of the pixels that are debiased (i.e., removed) $23.2\%$ are in the person and $76.8\%$ are in the background. Notably, this results in $20.4\% \pm 15.7\%$ of all the background pixels are masked out (see Appendix~\ref{sec:fair_blind_appendix} for more results). While fairness-through-blindness methods may purport to be removing gender cues primarily from the person, we find that much of the removed information comes from important parts of the image that may be necessary for the downstream task of object recognition.

%% file: Input/Discussion.tex
The results of most of our experiments are that the labelled gender expression is consistently discernible, i.e., classifiable at above random chance. This result holds even when an image is reduced to its average color (i.e. RGB values). Removal of such artifacts can be harmful if they lead to downstream allocational or representational harms (e.g., stereotyping, differences in  performance)~\cite{barocas2017problem,blodgett2020language}. 

\subsection{Implications for bias mitigation}
\iccvnew{Fairness-through-blindness approaches are still common in computer vision; the community has not fully shifted to the fairness-aware approach~\cite{pmlr-v81-dwork18a,lin2022fairgrape} this analysis recommends.} Prior work in computer vision fairness~\cite{bhargava2019caption,hendricks2018women,wang2019balanced}
has operated under the notion that gender bias in datasets solely comes from a person's appearance and not differences in the background or the average color of the image. Thus, these mitigation efforts attempt to remove gender bias by directly removing gender artifacts from the person. We engage with these prior works by showing that visual datasets have distinctly different distributions for images containing people of different genders. Similar to how proxies for protected attributes exist in tabular data~\cite{caton2020fairness,corbett2018measure,fisher2018all} thus rendering efforts to naively remove these attributes ineffective, removing gender artifacts in computer vision bias mitigation techniques may also be ineffective.

Overall, our results have three main implications for bias mitigation methods. First, gender cannot be removed from a visual dataset without removing information that is relevant for the downstream task (e.g., removing objects, scene). Thus, fairness-through-blindness approaches inevitably remove information that affects the model's performance. Second, while some methods assume removing part of a person’s appearance will remove the parts of the image that reveal gender~\cite{wang2019balanced}, we find in practice that these methods heavily rely on artifacts outside of the person. Finally, we encourage researchers and practitioners to shift away from fairness-through-blindness approaches towards those considering \emph{fairness through awareness}~\cite{dwork2011awareness}, as gender artifacts are simply too ubiquitous to consider trying to remain blind to. Existing fairness-aware methods first explicitly encode domain information and then explicitly mitigate the domain information~\cite{dwork2011awareness}. One form of fairness-aware methods can include constructing decoupled classifiers~\cite{pmlr-v81-dwork18a} in which a separate classifier is trained for different sub-groups of a sensitive attribute such as gender.
These algorithms do not try to erase gender, rather to accept its existence, and adjust algorithms accordingly.

\subsection{Implications for dataset construction} 
The solution is not to accept different distributions of images for people of different genders as an inevitable conclusion and make no efforts during dataset construction to consider gender representation. Rather, we point out these potentially harmful correlations and encourage practitioners to decide what kinds of gender artifacts they deem permissible or not in each specific context. 

The gender artifacts we discover and the differences in input distribution more broadly are not necessarily harmful. These differences can often represent historically important distinctions between groups of people. Moreover, gender, socially constructed as it is, is important to many social concerns about equal treatment~\cite{lorber2001gender,ridgeway2011framed,risman2018gender}. A participatory and iterative process can help to elucidate which artifacts should be removed (e.g., offensive stereotypes) and which may be appropriate to persist depending on the context. 

It is also important to consider the downstream task to determine what kinds of gender artifacts are permissible. One way of determining what kinds of artifacts may be more permissible is performing \emph{disaggregated measures of evaluation}. Of course, there are many nuances here to consider as well~\cite{barocas2021disagg}, and awareness about prominent gender artifacts can help guide the selection of axes along which to perform disaggregated analysis (beyond just demographic groups). For example, if scene is known to be a prominent gender artifact and the downstream task is scene classification, evaluation could be done across demographic group as well as scene type, to understand if specific scenes are always misclassified for people of a particular gender. Another way can be through engaging with communities affected by these downstream technologies~\cite{bennett2020point,blodgett2020language,katell2020toward}, as some gender effects such as those encoding harmful stereotypes, may be inherently harmful.

Finally, it is important to note we are only stating the \textit{presence} of such artifacts and not claiming these artifacts are necessarily harmful. While our analysis reveals these artifacts do exist in datasets, it does not guarantee the artifacts will be captured by a task-specific downstream problem or, even if they are captured, whether they lead to any harms. Nonetheless, it can be useful for practitioners to identify artifacts, following a similar process to that used in this work, and provide documentation for dataset users.

\subsection{Incoherence of gender predictions}
Finally, our analysis of gender artifacts elucidates another takeaway: we might consider that what our gender artifact model (i.e., a gender classifier) predicts in these contexts to be a nearly incoherent concept. The fact that the outline of a person, or even a pixelated 7x7 square, leads to high predictive performance of the notion of gender we have operationalized, should lead us to wonder about what exactly the model is predicting. Perhaps what the model is actually outputting is ``the gender expression of the person most likely to be holding an umbrella," or ``the gender expression of the person most likely to be in a running pose." All of these are notably different from predicting the person's gender.

While normative arguments against gender prediction ~\cite{keyes2018agr,larson-2017-gender,scheuerman2020} are sufficient to hinder its usage, our results suggest that a model's prediction of gender has little to do with the notion of gender the model designer is interested in. Although explicit gender prediction is now less common due to consciousness-raising on the topic, it is still prevalent in image captioning~\cite{bennett2021accessibility,bhargava2019caption}. 
As prior work~\cite{scheuerman2019computers} has shown, different gender classification systems frequently produce conflicting gender annotations. The concept of ``gender'' the model learns is as likely to be the most salient visual attribute correlated with gender in the dataset, as it is any kind of gender performance we deem socially meaningful~\cite{butler2002gender}. We must be careful should we ever find ourselves imbuing any meaning or value to the ``gender'' prediction of a model. 

%% file: Input/Conclusion.tex
Given the importance of datasets in computer vision, works critically examining these datasets provide insight into their limitations and have implications for the many methods evaluated on them. We go beyond prior works that analyze only annotated attributes and conduct a more comprehensive analysis of gender artifacts by studying what a ``gender classifier" relies on to make predictions. We analyze gender artifacts by occluding various combinations of color, person and background, and contextual objects. Gender is consistently discernible despite different manipulations to the input image. This analysis highlights the infeasibility of removing gender artifacts as a means for bias mitigation. Further, our work points to a general incoherence in the outputs of gender prediction models, as they appear to rely on any number of spurious correlations.

The permeation of gender artifacts throughout image datasets reveals it is futile to control for all possible differences. The aim for researchers should not be to remove these artifacts, but rather critically consider which gender artifacts are permissible and further to propose methods that are robust to these distribution differences. 

%% file: Input/Appendix.tex
In the supplementary material, we provide additional details on some sections of the main paper.

Sec.~\ref{sec:artifact_constraints_appendix}. Additional explanation for our constraint on artifact selection.

Sec.~\ref{sec:genderlabels_appendix}. Details on how gender labels are automatically derived.

Sec.~\ref{sec:hyperparam_appendix}. Additional gender artifact model training details.

Sec.~\ref{sec:contextual_obj_appendix}. Additional experiments regarding contextual objects described in Sec.~\ref{sec:contextualobjs} in the main paper.

Sec.~\ref{sec:person_scene_appendix}. Additional details on the person and background occlusions described in Sec.~\ref{sec:personscene} in the main paper.

Sec.~\ref{sec:balance_appendix}. Results from training and evaluating on a gender balanced dataset because the original datasets skewed male. 

Sec.~\ref{sec:fair_blind_appendix}. Additional results on our fairness-through-blindness case study.

\setcounter{figure}{5}
\section{Constraints on artifact selection}
\label{sec:artifact_constraints_appendix}
We analyze gender artifacts that are learnable (i.e., result in a learnable difference for a machine learning model) and interpretable (i.e., have an interpretable human corollary). We implement the constraint that the gender artifacts must be interpretable as artifacts perceptible to humans often have the most pressing fairness concerns (e.g., an imperceptible artifact such as a correlation between the n-th pixel in the image and gender may not have as high pressing fairness concerns as if there was a high correlation between outdoor backgrounds and male gender labels). However, we acknowledge there can be infinitely many potential correlations in an image and this criterion is non-exhaustive. 

\section{Automatically deriving gender labels}
\label{sec:genderlabels_appendix}
Following prior work~\cite{wang2019balanced,mals}, we use the captions from the Common Objects in Context (COCO) to derive gender labels (Sec.~\ref{sec:method}). Concretely, we first convert the captions to lowercase. Then, using the following gendered set of words from Zhao \textit{et al.}~\cite{zhao2021captionbias}, we query our captions for the presence of these keywords: [``male,'' ``boy,'' ``man,'' ``gentleman,'' ``boys,'' ``men,'' ``males,'' ``gentlemen''] and [``female", ``girl", ``woman", ``lady", ``girls", ``women", ``females", ``ladies"]. We assign the respective gender label if two of the five captions contain a gendered keyword and discard images for which the captions contain both male and female keywords. We choose to use two of the five captions as these automatically derived captions match the explicit annotations from labelers $85.4\%$ of the time~\cite{zhao2021captionbias}. This suggests label noise for gender labels in the COCO dataset does not significantly affect our results.

\section{Additional model training details}
\label{sec:hyperparam_appendix}
\smallsec{Gender artifact model}
 The model is optimized with stochastic gradient descent (SGD) with a momentum of $0.9$ and a batch size of 64. The ResNet-50 has a final fully connected layer mapping the 2048 size hidden layer to a single output value, and we arbitrarily assign ``male'' as equal to 0 and ``female'' to 1. We optimize the hyperparameters for each model on the validation set using grid search (learning rate: $\{10^{-2},10^{-3},\dots10^{-5}\}$; weight decay: $\{10^{-2},10^{-3},\dots10^{-5}\}$). For the baseline model, input images are resized to $224$ x $224$ and randomly flipped horizontally during training. Furthermore, we bootstrap until convergence (5,000 resamples) and report a $95\%$ confidence interval on the test set. 

\begin{figure}[t!]
\centering
\includegraphics[width=\linewidth]{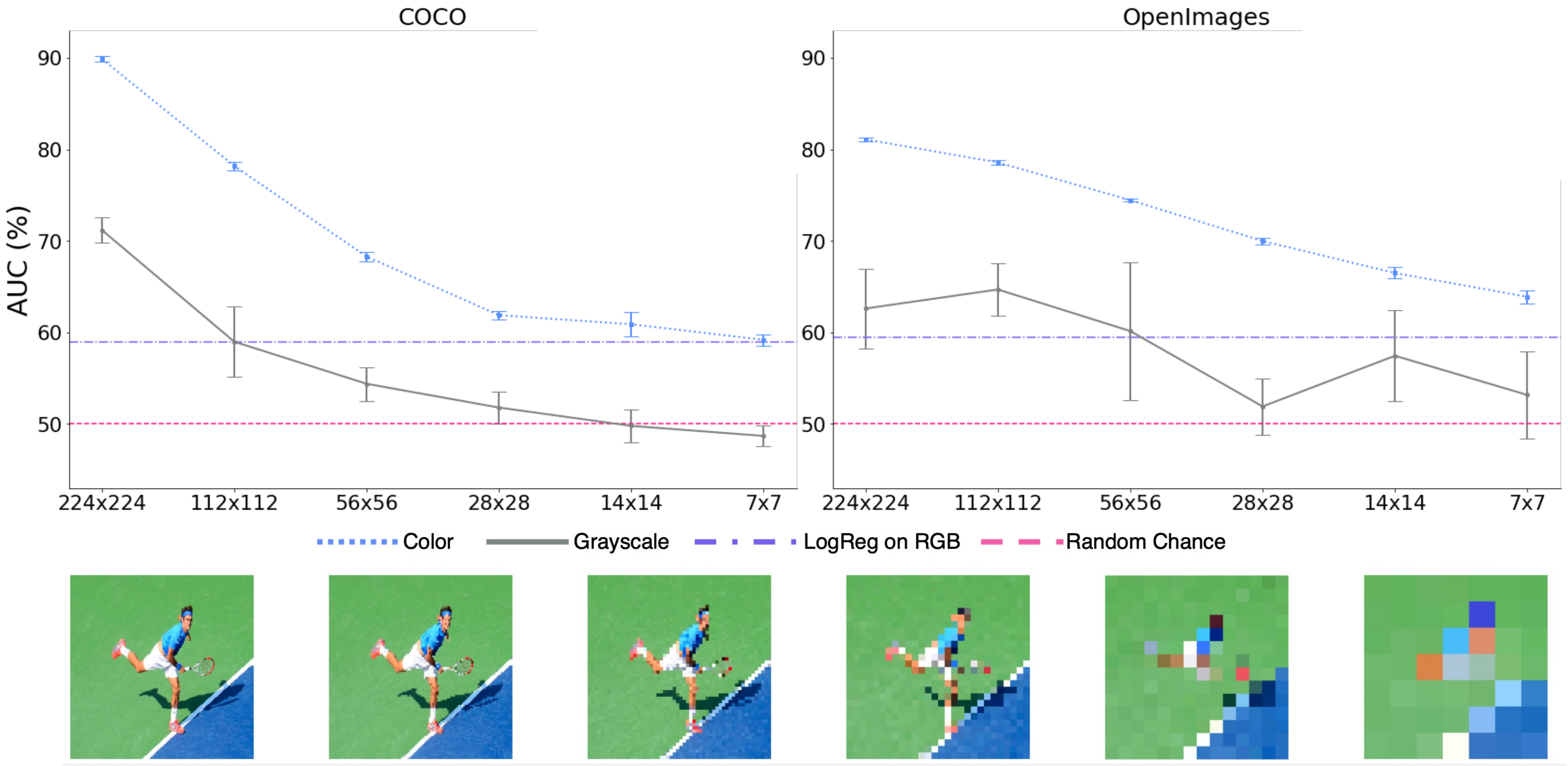}
\caption{\textbf{AUC of models trained with varying resolution and color of input
images.} 
We display another method of calculating confidence intervals for the ResNet-50 model by training five models on random seeds and provide the 95\% confidence
interval.}
\label{fig:color_appendix}
\end{figure}

\smallsec{Training models on random seeds}
We provide the 95\% confidence intervals through bootstrapping until convergence (5,000 resamples) for all of results. As an alternative means for providing confidence intervals, for all ResNet-50 models, we train five models on random seeds and provide the 95\% confidence intervals as well. As seen in Tbl.~\ref{tab:randomseeds}, ~\ref{tab:pretraining}, and Fig.~\ref{fig:color_appendix} the intervals are similar to those found via bootstrapping.   When removing random contextual objects, we also report the result of training five separate random classifiers as displayed in Fig.~\ref{logreg_removeobjs_appendix}. 
In all cases, the gender artifact model continues to perform above random chance. 

\begin{table}[t!]
    \footnotesize
    \centering
    \begin{tabular}{p{3cm}p{2cm}p{2cm}}
    \toprule
         &  COCO & OpenImages \\
    \hline
        Full & $93.4 \pm 0.1$ & $81.2 \pm 0.2$\\
        Full NoBg & $92.6 \pm 0.1$ & -\\
        MaskSegm & $79.5 \pm 0.2$ & -\\
        MaskRect & $70.7 \pm 0.1$ &  $63.3 \pm 0.4$\\
        MaskSegm NoBg & $76.0 \pm 0.1$ & -\\
        MaskRect NoBg & $58.3 \pm 0.2$ & $59.7 \pm 0.7$\\
    \bottomrule
    \end{tabular}
    \caption{\textbf{Performance of model trained using five random seeds.} We report the mean AUC (\%) and a 95\% confidence interval of five gender artifact models trained using random seeds.}
    \label{tab:randomseeds}
\end{table}

\smallsec{Pre-training on different datasets}\label{sec:pretrain_appendix}
In addition to pre-training on ImageNet as in Sec.~\ref{sec:personscene}, we evaluate a ResNet-50 pre-trained on Places-365~\cite{zhou2017places} and PASS MoCo-v2~\cite{asano21pass} weights. We do not train a model from scratch as there is insufficient data. All models perform above random chance (Tbl.~\ref{tab:pretraining}). In particular, the PASS dataset does not contain any people, indicating that the difference observed is attributable to differences in the input distributions, not the dataset on which the model was pre-trained.
\begin{table}[t!]
    \footnotesize
    \centering
    \begin{tabular}{p{2.5cm}p{1.5cm}p{1.5cm}p{1.5cm}}
    \toprule
    & ImageNet & Places-365 & PASS \\
    \midrule
    Full & $93.4 \pm 0.2$ & $88.4 \pm 0.2$ & $89.9 \pm 0.2$\\ 
    Full NoBg & $92.7 \pm 0.2$ & $88.5 \pm 0.2$ & $89.9 \pm 0.2$\\
    MaskSegm & $79.6 \pm 0.3$ & $74.6 \pm 0.3$ & $75.8 \pm 0.3$\\
    MaskRect & $70.8 \pm 0.3$ & $69.5 \pm 0.4$ & $68.6 \pm 0.4$\\
    MaskSegm NoBg & $74.8 \pm 0.3$ & $70.9 \pm 0.4$ & $74.1 \pm 0.3$\\
    MaskRect NoBg & $58.0 \pm 0.4$ & $58.4 \pm 0.4$ & $58.8 \pm 0.4$\\ 
    \bottomrule
    \end{tabular}
    \caption{\textbf{Pre-training on different datasets.} We report the AUC (\%) of a gender artifact model that is pre-trained on one of three datasets: ImageNet~\cite{russakovsky2015ilsvrc}, Places-365~\cite{zhou2017places}, and PASS~\cite{asano21pass}. The model is then trained and evaluated on different occlusions (introduced in Sec.~\ref{sec:personscene}) on the COCO dataset.}
    \label{tab:pretraining}
\end{table}

\begin{table}[t!]
    \footnotesize
    \centering
    \begin{tabular}{ll}
         \midrule
        Male & Female\\
        \hline
         Trumpet & Break\\
         Weapon & Tart \\
         Bathtub & Shotgun \\
         Billboard & Dairy Product\\
         Sombrero & Goat\\
         Tiara & Ambulance\\
         Ceiling Fan & Duck\\
         Scoreboard & Banana\\
         Missile & High Heels\\
         Cupboard & Bow and Arrow\\

         \bottomrule
    \end{tabular}
    \caption{\textbf{Relevant contextual objects.} The ten most relevant objects in descending order, as identified by the weights of the logistic regression classifier trained on OpenImages.}
    \label{tab:oi_contextualobjects}
\end{table}

\section{Additional contextual object experiments}
\label{sec:contextual_obj_appendix}
\smallsec{Additional contextual object results}
For COCO, in addition to reporting the results from bootstrapping, we report the results from training five separate random classifiers and report the standard deviation. See Fig.~\ref{logreg_removeobjs_appendix} for results.

Next, for OpenImages, in Tbl~\ref{tab:oi_contextualobjects}, we report the 10 most relevant objects in descending order, as identified by the weights of the logistic regression classifier trained on OpenImages.

\begin{figure}[t!]
\centering
\includegraphics[width=\linewidth]{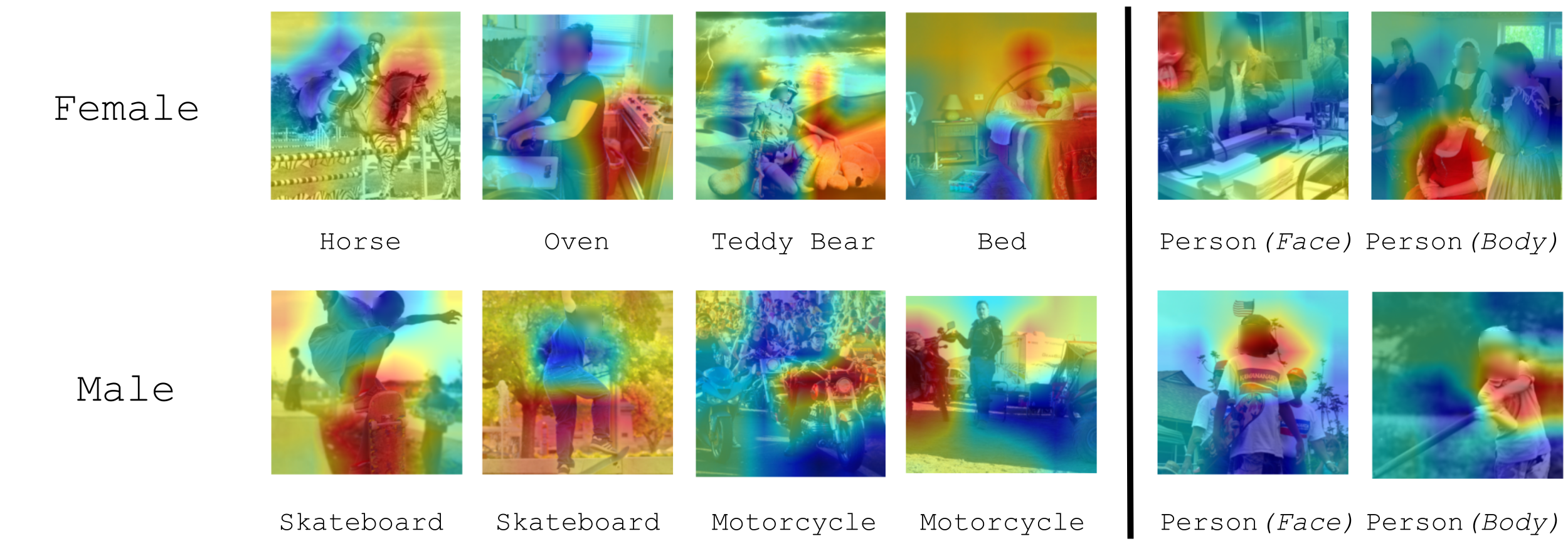}
\caption{\textbf{Visualizing model attention.} On the left four images, the gender artifact model's attention is on contextual objects and not the person, suggesting the model relies on spurious correlations to infer gender. On the right, the model's attention is on the person. These qualitative CAM analyses suggest gender artifacts are embedded in the background of images (i.e., beyond the person) as observed by Hendricks \textit{et al.}~\cite{hendricks2018women} and motivate future analysis in contextual objects. 
}
\label{fig:CAM_top10}
\end{figure}

\smallsec{Visualizing gender artifacts through model attention}
To better understand \textit{what} the model relies on in the background, we visualize Class Activation Maps (CAMs) \cite{selvaraju2017gradcam} to see which contextual objects the model's attention focuses on. CAMs are saliency maps shown to expose the implicit attention of neural networks, highlighting “the most informative image regions relevant to the predicted class.” \cite{selvaraju2017gradcam}
For example, the model tends to focus on various spurious correlations such as indoor objects (\texttt{oven} and \texttt{bed}) to classify female and outdoor objects (\texttt{skateboard} and \texttt{motorcycle}) to classify male (Fig.~\ref{fig:CAM_top10}).

\smallsec{Performance of contextual objects in the logistic regression model}
In addition to reporting the results of bootstrapping in the main paper, we also report the results of running five separate random classifiers and report the standard deviation for the model where we remove random objects (Fig.~\ref{logreg_removeobjs_appendix}).
\begin{figure}[]
\centering
\includegraphics[width=\linewidth]{Figures/fig7_appendix.pdf}
\caption{\textbf{Performance of contextual objects in the Logistic Regression Model (COCO).} 
We visualize the change in AUC as objects are iteratively removed from the object-based logistic regression classifier to see how many objects are required before the classifier performs at random chance. When removing random objects, we train five separate random classifiers and report the standard deviation. We report the most relevant objects in descending order, as identified by the weights of a logistic regression classifier trained for gender prediction.
}
\label{logreg_removeobjs_appendix}
\end{figure}

\section{Person and background occlusions} 
\label{sec:person_scene_appendix}
\begin{figure}[]
    \centering
    \includegraphics[width=\linewidth]{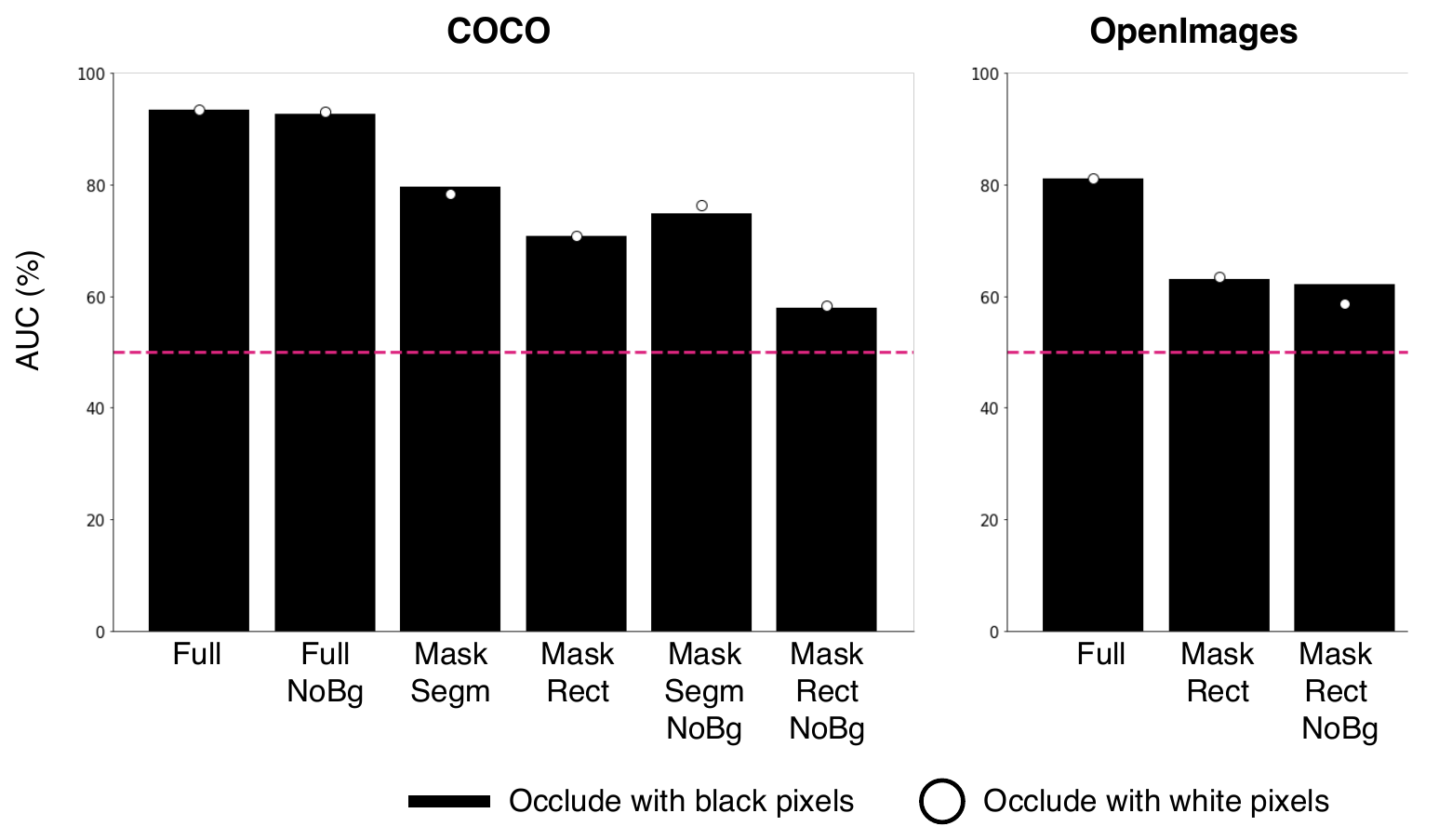}
    \caption{\textbf{Comparing AUCs when occluding with different color pixels.} We report the AUC (\%) of the gender artifact model when we occlude using black pixels (also reported in Sec.~\ref{sec:personscene}) in the black bars and when we occlude using white pixels in the white points. The pink line represent the AUC at random chance.}
    \label{fig:white_occ}
\end{figure}

\smallsec{Robustness to occlusion color}
In Sec.~\ref{sec:personscene}, we occlude person and background cues using black pixels. To demonstrate that occlusions are robust to the color of the pixels, we present the AUCs when we occlude using white pixels instead (see Fig.~\ref{fig:white_occ}). As seen in Tab.~\ref{tab:personscene}, the model performance does not change considerably when we use white pixels as opposed to black pixels for our occlusions. The largest change in AUC for COCO is $1.5\%$ on MaskSegm NoBg and for OpenImages is $3.5\%$ for MaskRect NoBg. Further, the ranking of AUCs across occlusions does not change. 

\begin{table}[t!]
    \footnotesize
    \centering
    \begin{tabular}{p{2cm}p{1.2cm}p{1.2cm}|p{1.2cm}p{1.2cm}}
    \toprule
         &  \multicolumn{2}{c|}{COCO} & \multicolumn{2}{c}{OpenImages}\\
         &  B & W & B & W \\
    \hline
        Full & $93.4 \pm 0.2$ & $93.4 \pm 0.2$ & $81.1 \pm 0.3$ & $81.1 \pm 0.3$\\
        Full NoBg & $92.7 \pm 0.2$ & $93.0 \pm 0.2$  & - & -\\ 
        MaskSegm & $79.6 \pm 0.3$ & $78.4 \pm 0.3$ &-& -\\
        MaskRect & $70.8 \pm 0.3$ & $70.7 \pm 0.3$ & $63.1 \pm 0.4$ & $63.5 \pm 0.4$ \\
        MaskSegm NoBg & $74.8 \pm 0.3$ & $76.3 \pm 0.3$  &-& -\\
        MaskRect NoBg & $58.0 \pm 0.4$ & $58.3 \pm 0.4$ & $62.2 \pm 0.4$ & $58.7 \pm 0.4$ \\
    \bottomrule
    \end{tabular}
    \caption{\textbf{Performance of gender artifact model on various occlusions.} We report the AUC (\%) of the gender artifact model on various occlusions when we use black pixels (B) versus white pixels (W) for both COCO~\cite{lin2014microsoft} and OpenImages~\cite{kuznetsova2020open}.}
    \label{tab:personscene}
\end{table}

\smallsec{Additional experimental settings}
In addition to the six settings we present in Sec.~\ref{sec:personscene}, we analyze three more settings. In the first, we occlude the background around the bounding box of the person. This achieves an AUC of $93.0 \pm 0.2$ and $89.6 \pm 0.2$ for COCO and OpenImages respectively. Next, we occlude only the person's face which was detected using Amazon Rekognition and MTCNN~\cite{zhang2016joint}. This yields an AUC of $92.0 \pm 0.2$ and $79.8 \pm 0.3$. Finally, only on the COCO images, we occlude the background and include only the person's skeleton, which achieves an AUC of $65.6 \pm 0.4$.

\begin{figure}
    \centering
\includegraphics[width=\linewidth]{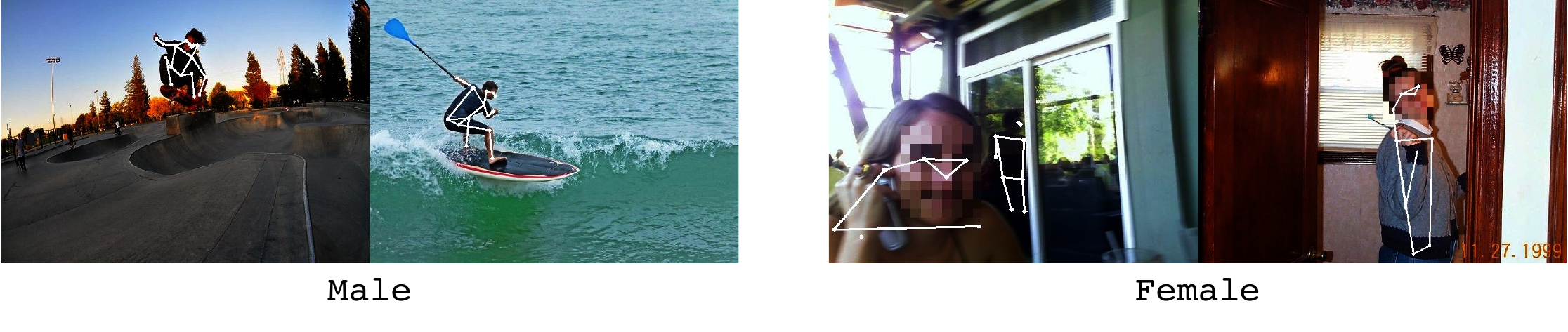}
    \caption{\textbf{Qualitative examples of differences in poses.} We visualize poses, overlayed on the original COCO image, that were predicted as highly male (top) and highly female (bottom). Face pixelation is not in the COCO image but included in an effort to partially preserve privacy.}
    \label{fig:poses}
\end{figure}

\smallsec{Pose analysis} After training a model just on the person's keypoints from COCO, we qualitatively inspect the poses with the greatest absolute scores. In Fig. 4, we display the poses predicted to be more likely to be male and female and notice that images predicted to. be male are smaller and in action (e.g., playing a sport, jumping) whereas those predicted to be more likely to be female tended to be larger and standing still.

\section{Results on a balanced dataset}
\label{sec:balance_appendix}
Both COCO and OpenImages are skewed male (69.2\% and 61.1\% of the training set for COCO and OpenImages). We examine whether the gender imbalance in the dataset affects the discoverable artifacts. Concretely, we train and evaluate our gender artifact model on a balanced dataset. As shown in Tbl.~\ref{tab:balanced}, the AUCs are comparable to the unbalanced dataset. In fact, the largest change of AUC was from 65.6 on the skewed dataset to 63.6 on the balanced for NoPix-SomeSh-NoBg. 

\begin{table}[]
    \footnotesize
    \centering
    \begin{tabular}{rrr}
    \toprule
         & Original & Balanced \\
    \hline
        Pix- Sh-Bg & 93.4 & 93.3 \\
        Pix-Sh-SomeBg & 93.0 & 93.0\\
        Pix-Sh-NoBg & 92.6 & 92.6 \\ 
        NoFace-Sh-Bg & 92.0 & 91.7 \\ 
        NoPix-Sh-Bg & 79.4 & 78.1\\
        NoPix-NoSh-Bg & 70.7 & 69.4 \\
        NoPix-Sh-NoBg & 76.0 & 76.7  \\
        NoPix-SomeSh-NoBg & 65.6 & 63.6\\
        NoPix-NoSh-NoBg & 58.4 & 58.9\\
    \bottomrule
    \end{tabular}
    \caption{\textbf{Results on gender-balanced datasets} We report the AUC (\%) of the model trained on a gender-balanced dataset}
    \label{tab:balanced}
\end{table}

\section{Additional fairness-through-blindness results}
\label{sec:fair_blind_appendix}
\begin{figure}[t!]
    \centering
    \includegraphics[width=0.75\linewidth]{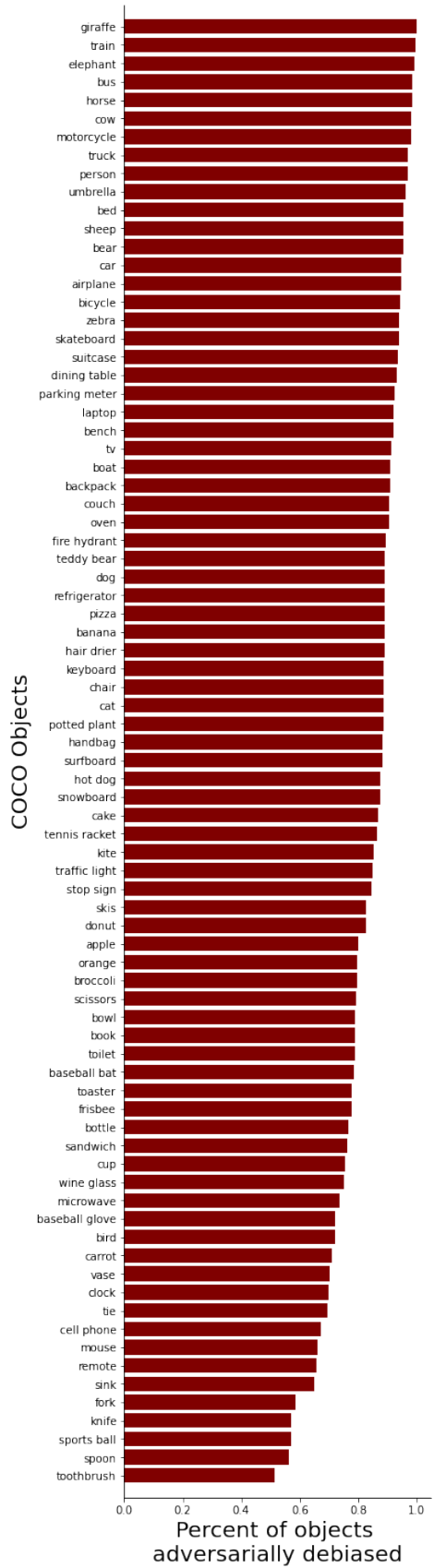}
    \caption{\textbf{Occluded contextual objects during adversarial debiasing.} We visualize the additional results from our fairness-through-blindness case study in which we analyze the COCO objects that were most likely adversarially debiased (i.e., contribute to gender information). }
    \label{fig:fairblind_appendix}
\end{figure}

We provide additional results on the COCO objects that were most likely to be affected by the adversarial de-biasing method. The top five non-\texttt{person} objects, which we calculate as the ratio of images in which the object is affected to the number of images that object occurs in, are as follows: \texttt{giraffe}, \texttt{train}, \texttt{elephant}, \texttt{bus}, and \texttt{horse}. It is likely that these objects are more likely to be occluded as a person may be directly interacting with the object (e.g., riding a \texttt{horse}) in the image. We provide the full COCO object results in Fig.~\ref{fig:fairblind_appendix}.

